\documentclass[letterpaper, 10 pt, conference]{ieeeconf}  
\IEEEoverridecommandlockouts
\usepackage{times}
\usepackage{xcolor}
\usepackage{multicol}
\usepackage{multirow}
\usepackage{breqn}
\usepackage[bookmarks=true]{hyperref}
\usepackage{amsmath,amssymb,accents,mathtools,amsfonts}

\usepackage{amsthm}
\usepackage{algorithm}
\usepackage{algorithmic}
\usepackage{makecell}
\usepackage[font=footnotesize]{caption}
\usepackage{subcaption}
\usepackage[bb=boondox]{mathalfa}
\usepackage{graphicx}

\usepackage[
   style=ieee,
    doi=false,
    isbn=false,
    url=true,
    eprint=false,
    backend=biber,
    natbib=true
    ]{biblatex}
    
\bibliography{Gait}

\begin{document}
\providecommand{\R}{\ensuremath \mathbb{R}}
\providecommand{\N}{\ensuremath \mathbb{N}}

\newtheorem{defn}{Definition}
\newtheorem{rem}[defn]{Remark}
\newtheorem{lem}[defn]{Lemma}
\newtheorem{assum}[defn]{Assumption}
\newtheorem{ex}[defn]{Example}
\newtheorem{thm}[defn]{Theorem}
\newtheorem{cor}[defn]{Corollary}

\newcommand{\norm}[1]{\left\Vert#1\right\Vert}
\newcommand{\abs}[1]{\left\vert#1\right\vert}

\providecommand{\PP}{\mathcal{P}}
\providecommand{\LL}{\mathcal{L}}
\providecommand{\W}{\mathcal{W}}
\providecommand{\D}{\mathcal{D}}
\providecommand{\x}{\mathbf{x}}
\providecommand{\X}{\mathcal{X}}
\providecommand{\K}{\mathcal{K}}
\providecommand{\ZZ}{\mathcal{Z}}
\providecommand{\XY}{\mathcal{XY}}
\renewcommand{\P}{\mathcal{P}}
\providecommand{\G}{\mathcal{G}}
\providecommand{\B}{\mathcal{B}}
\providecommand{\HH}{\mathcal{H}}
\providecommand{\Z}{\mathcal{Z}}
\providecommand{\A}{\mathcal{A}}
\providecommand{\V}{\mathcal{V}}
\providecommand{\U}{\mathcal{U}}
\providecommand{\T}{\mathcal{T}}
\providecommand{\J}{\mathcal{J}}
\providecommand{\Y}{\mathcal{Y}}
\providecommand{\RR}{\mathcal{R}}
\providecommand{\Q}{\mathcal{Q}}
\providecommand{\HH}{\mathcal{H}}
\providecommand{\I}{\mathcal{I}}
\providecommand{\E}{\mathcal{E}}
\providecommand{\F}{\mathcal{F}}
\renewcommand{\SS}{\mathcal{S}}
\providecommand{\W}{\mathcal{W}}
\providecommand{\OO}{\mathcal{O}}

\providecommand{\Tc}{\Tilde{c}}
\providecommand{\Ta}{\Tilde{\alpha}}
\providecommand{\TL}{\Tilde{\mathcal{L}}}

\providecommand{\Wplus}{\W^{+}}

\providecommand{\intG}{[\mathcal{G}]}
\providecommand{\intdG}{[\dot{\mathcal{G}}]}

\newcommand{\OPToffline}{\textnormal{\texttt{OPT-offline}}}
\newcommand{\OPTonline}{\textnormal{\texttt{OPT-online}}}

\newcommand{\ubar}[1]{\underaccent{\bar}{#1}}

\providecommand{\qA}{q_A}
\providecommand{\qj}{q_j}
\providecommand{\qu}{q_u}
\providecommand{\qa}{q_a}

\providecommand{\dq}{\dot{q}}
\providecommand{\dqj}{\dot{q}_j}
\providecommand{\dqu}{\dot{q}_u}
\providecommand{\dqa}{\dot{q}_a}
\providecommand{\ddq}{\ddot{q}}
\providecommand{\ddqj}{\ddot{q}_j}
\providecommand{\ddqu}{\ddot{q}_u}
\providecommand{\ddqa}{\ddot{q}_a}

\providecommand{\dqm}{\dot{q}_{ma}}
\providecommand{\ddqm}{\ddot{q}_{ma}}

\providecommand{\qd}{q_d}
\providecommand{\qdj}{q_{dj}}
\providecommand{\qda}{q_{da}}
\providecommand{\dqd}{\dot{q}_d}
\providecommand{\dqdj}{\dot{q}_{dj}}
\providecommand{\dqda}{\dot{q}_{da}}
\providecommand{\ddqd}{\ddot{q}_d}
\providecommand{\ddqdj}{\ddot{q}_{dj}}
\providecommand{\ddqda}{\ddot{q}_{da}}

\providecommand{\qGait}{q}
\providecommand{\qGaitj}{q_{j}}
\providecommand{\qGaita}{q_{a}}
\providecommand{\qGaitjzero}{q_{j0}}
\providecommand{\dqGait}{\dot{q}}
\providecommand{\dqGaitj}{\dot{q}_{j}}
\providecommand{\dqGaita}{\dot{q}_{a}}
\providecommand{\dqGaitjzero}{\dot{q}_{j0}}
\providecommand{\ddqGait}{\ddot{q}}
\providecommand{\ddqGaitj}{\ddot{q}_{j}}
\providecommand{\ddqGaita}{\ddot{q}_{a}}
\providecommand{\ddqGaitjzero}{\ddot{q}_{j0}}

\newcommand{\wdistlong}{w(\Delta)}
\newcommand{\wdist}{w}
\newcommand{\wdistlongi}{w_i(\Delta)}
\newcommand{\wdistinterval}{w}

\providecommand{\pst}{p_{st}}
\providecommand{\tst}{\theta_{st}}
\providecommand{\psw}{p_{sw}}
\providecommand{\asw}{\alpha_{sw}}
\providecommand{\pswx}{p_{sw,x}}
\providecommand{\pswy}{p_{sw,y}}
\providecommand{\pswz}{p_{sw,z}}
\providecommand{\tp}{\tilde{p}}
\providecommand{\tpsw}{\tilde{p}_{sw}}
\providecommand{\tpj}{\tilde{p}_{j}}
\providecommand{\tpjx}{\tilde{p}_{j,x}}
\providecommand{\tpjy}{\tilde{p}_{j,y}}
\providecommand{\tpjz}{\tilde{p}_{j,z}}
\providecommand{\ta}{\tilde{\alpha}}
\providecommand{\tasw}{\tilde{\alpha}_{sw}}
\providecommand{\taj}{\tilde{\alpha}_{j}}

\newcommand{\red}[1]{{\color{red} #1}}
\newcommand{\blue}[1]{{\color{blue} #1}}

\providecommand{\GaitLibOne}{\texttt{Gaits-One}}
\providecommand{\GaitLibTurn}{\texttt{Gaits-Turn}}
\providecommand{\GaitLibSix}{\texttt{Gaits-Six}}

\providecommand{\FRS}{FRS}
\providecommand{\FRSz}{\ensuremath FRS_\text{hip}}
\providecommand{\FRShipL}{\ensuremath FRS_{\text{\normalfont hip},L}}
\providecommand{\FRShipR}{\ensuremath FRS_{\text{\normalfont hip},R}}
\providecommand{\FRSxy}{\ensuremath FRS_\text{\normalfont pelvis}}
\providecommand{\zhip}{\ensuremath z_\text{hip}}
\providecommand{\zhipdot}{\ensuremath \dot{z}_\text{hip}}
\providecommand{\Tplan}{\ensuremath \tau_\text{plan}}
\providecommand{\Tcommand}{\ensuremath T_e}
\providecommand{\Tslot}{\ensuremath T_s}
\providecommand{\Tprocess}{\ensuremath T_c}
\providecommand{\Tsense}{\ensuremath T\sense}
\providecommand{\Ttrajopt}{\ensuremath T_\text{trajopt}}
\providecommand{\Tonline}{\ensuremath T_\text{plan}}
\providecommand{\Trrt}{\ensuremath T_\text{route}}
\providecommand{\Bl}{\ensuremath \underline{B}}
\providecommand{\Bu}{\ensuremath \overline{B}}
\providecommand{\FRSzOpt}{\textup{FRShipOpt}}
\providecommand{\FRSxyOpt}{\textup{FRSxyOpt}}
\providecommand{\TrajOpt}{\textup{\texttt{TrajOpt}}}
\providecommand{\Opt}{\textup{\texttt{Opt}}}
\providecommand{\SolveTrajOpt}{\textup{\texttt{SolveTrajOpt}}}
\newcommand{\obs}{_\text{obs}}
\newcommand{\plan}{_\text{plan}}
\newcommand{\sense}{_\text{sense}}
\newcommand{\hip}{_\text{hip}}
\newcommand{\pred}{_\text{pred}}
\newcommand{\goal}{_\text{goal}}
\providecommand{\hipzL}{_{\text{\normalfont hip},zL}}
\providecommand{\hipzR}{_{\text{\normalfont hip},zR}}
\providecommand{\hip}{_\text{\normalfont hip}}
\providecommand{\db}{\delta_b}
\providecommand{\ds}{\delta_s}
\providecommand{\M}{\mathcal{M}}

\newcommand{\trunc}{^\text{trunc}}
\newcommand{\propa}{^\text{prop}}
\newcommand{\ts}[1]{\textsuperscript{#1}}
\newcommand{\conv}{\text{conv}}
\newcommand{\BV}{\text{BV}}

\providecommand{\x}{\mathbf{x}}
\renewcommand{\L}{\mathcal{L}}
\providecommand{\parammap}{\ensuremath M}
\providecommand{\buffer}{\textup{\texttt{buffer}}}
\providecommand{\sample}{\textup{\texttt{sample}}}
\providecommand{\disc}{\texttt{disc}}

\providecommand{\zthreshold}{0.75}

\providecommand{\BRTD}{\text{BipedRTD}}
\newcommand{\hi}{_\text{hi}}
\providecommand{\tfin}{t_\text{f}}
\providecommand{\T}{\ensuremath T}
\newcommand{\defemph}[1]{\emph{#1}}

\newcommand{\iv}[1]{[ #1 ]}

\title{\LARGE \bf Rapid and Robust Trajectory Optimization for Humanoids}
\author{Bohao Zhang$^1$ and Ram Vasudevan$^2$
\thanks{$^{1}$Robotics Institute, University of Michigan, Ann Arbor, MI \texttt{jimzhang@umich.edu}.}
\thanks{$^{2}$Mechanical Engineering, University of Michigan, Ann Arbor, MI \texttt{ramv@umich.edu}.}
}

\maketitle
            
\begin{abstract}
    \label{sec:abstract}
Performing trajectory design for humanoid robots with high degrees of freedom is computationally challenging.
The trajectory design process also often involves carefully selecting various hyperparameters and requires a good initial guess which can further complicate the development process.
This work introduces a generalized gait optimization framework that directly generates smooth and physically feasible trajectories. 
The proposed method demonstrates faster and more robust convergence than existing techniques and explicitly incorporates closed-loop kinematic constraints that appear in many modern humanoids.
The method is implemented as an open-source C++ codebase which can be found at \href{https://roahmlab.github.io/RAPTOR/}{https://roahmlab.github.io/RAPTOR/}.
\end{abstract}
\section{Introduction}
\label{sec:introduction}

Numerically constructing energy efficient trajectories that accomplish a user-specified task for high degree of freedom robots is an important challenge.
A classic approach to perform this type of synthesis for bipeds has relied on representing the problem as a nonlinear optimization program.
These methods usually take several minutes to compute an optimal trajectory for bipeds (e.g., Cassie or Digit) \cite{paper-trajopt, paper-frost}.
Notably many modern bipeds have closed-loop kinematic constraints that require incorporating nonlinear constraints during optimization, which can make it even more computationally taxing to construct a trajectory via optimization.
Solving this nonlinear program often requires managing numerous hyperparameters that can significantly affect the performance of the optimization process.
Notably solving these nonlinear optimization problems often requires a good initial guess.

To improve the speed at which trajectory synthesis occurs, many researchers have simplified model complexity by utilizing reduced-order dynamic models \cite{paper-related-LIP0,paper-related-LIP1, paper-related-LIP2, paper-related-obstacle-avoidance, liu2020leveraging}.
Of particular note is the centroidal dynamics representation of a legged robot when performing trajectory optimization \cite{paper-centriodal}.
It is able to more tractably deal with kinematic obstacle avoidance constraints, but because it utilizes a reduced-order dynamic model during trajectory optimization the solution that it finds may not be directly realizable on a real legged system due to torque limits \cite[Section IV.A]{paper-optimization-overview}. 

\begin{figure}[t]
    \centering
    \includegraphics[width=0.75\columnwidth]{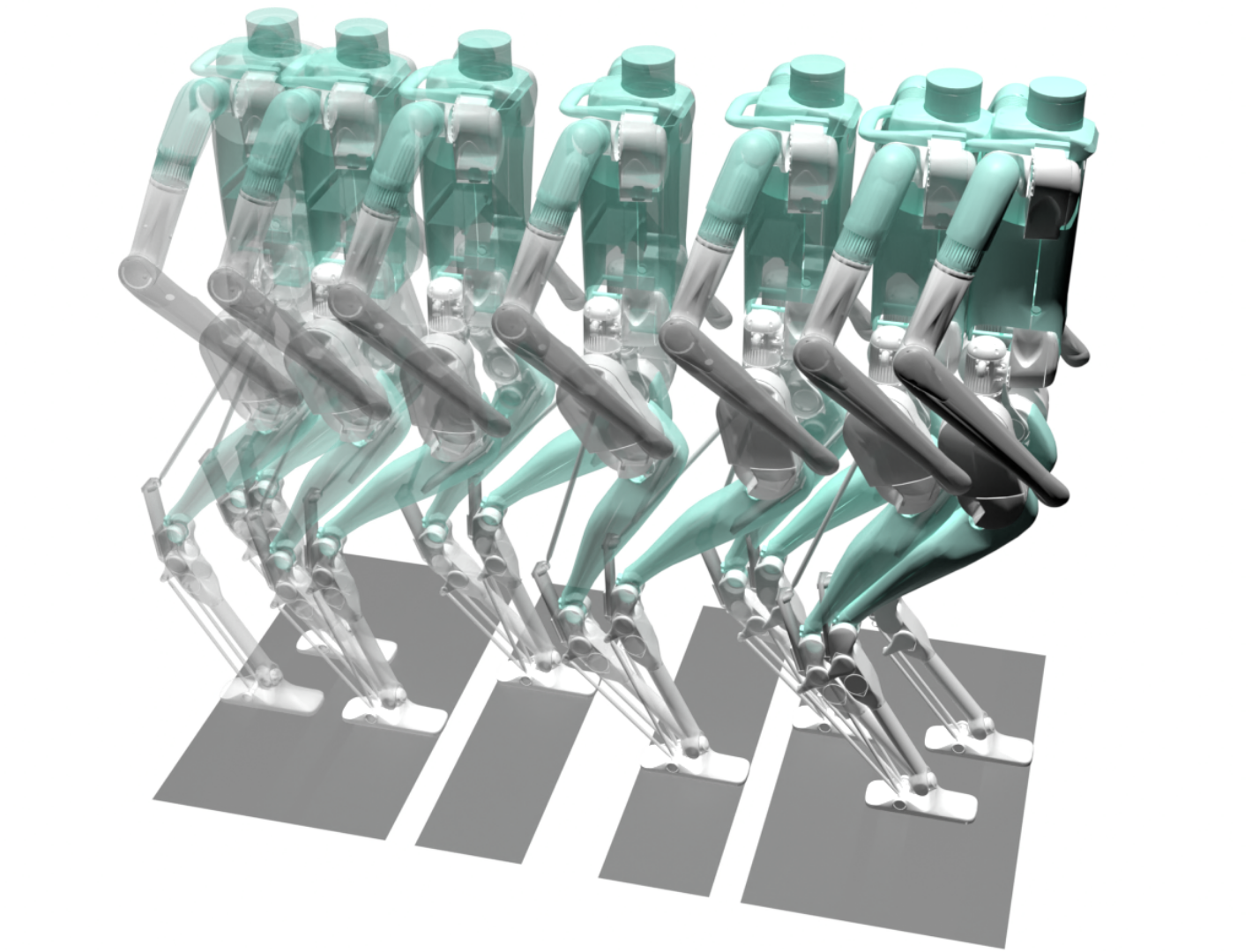}
    \caption{This figure illustrates a physically feasible six-step periodic gait with different step lengths for humanoid robot Digit that is found by RAPTOR, the trajectory optimization algorithm developed in this paper. 
    The duration of each step is fixed to be $0.4$ seconds, which yields a total duration of $2.4$ seconds of the whole gait.
    It only takes RAPTOR $16$ seconds to converge to a feasible solution.}
    \label{fig:summary}
\end{figure}

Rather than utilize a reduced order model, others have tried to improve the speed of computation of these nonlinear programs by reformulating the trajectory optimization problem.
For example,  researchers have utilized Differential Dynamic Programming (DDP) to perform energy efficient trajectory design for high degree of freedom robotic systemsin seconds \cite{paper-crocoddyl, paper-crocoddyl-inverse}.
However, the contact constraints or other kinematics constraints are treated as soft constraints that are directly added to the cost function to minimize.
This means that these kinematic constraints may not be perfectly satisfied at an optimal solution.
To address this limitation, researchers have recently illustrated how to perform DDP with rigid constraints \cite{paper-aligator}.
This approach is able to generate a solution that respects the system dynamics while satisfying other state or input constraints, but is slower than \cite{paper-crocoddyl}.

The emphasis of this work is on performing trajectory optimization when a user pre-specifies the contact sequence. 
However, there are methods that try to optimize over the contact sequence as well \cite{paper-idto,zhao2019optimal}, but because this dramatically increases the number of optimization variables it can be difficult to apply these approaches while performing gait optimization for humanoids. 

More recent work has applied deep learning to perform trajectory design for bipeds \cite{paper-related-learning1, paper-related-learning5}.
Using iterative reinforcement learning, researchers have learned a policy that can enable a robot to successfully traverse a variety of terrains, including slopes, stairs, or even blocks \cite{paper-related-learning4}.
These methods usually begin from or rely on an existing library of reference trajectories for the deep neural network to imitate \cite{paper-humanoid-transformer}.
This can be nontrivial to automatically generate on a high-dimensional robot.

This paper proposes a new gait optimization framework, RAPTOR, that optimizes over only the actuated joints of a robot and then uses inverse dynamics to reconstruct rest of the states.
This results in fewer decision variables and constraints during optimization-based trajectory design.
As a result, RAPTOR can perform trajectory design faster for modern bipeds which usually incorporate closed-loop kinematic constraints.
The contributions of this work are three-fold:
First, in Section \ref{sec:gaitgeneration}, a new formulation of a generalized gait optimization framework for dynamic bipedal locomotion that directly generates a smooth trajectory that is physically feasible and can incorporate closed-loop constraints.
Second, in Section \ref{sec:simulation}, we demonstrate that our method can be effectively scaled to a high-dimensional full-size humanoid and achieve faster and more robust convergence compared with other methods.
Finally, we release the code-base, which is written entirely in C++, as an open source package \href{https://github.com/roahmlab/RAPTOR}{https://github.com/roahmlab/RAPTOR}.

\section{Preliminary}
\label{sec:preliminary}

This section formulates the bipedal robot dynamics as well as certain assumptions that are used to formulate a trajectory optimization problem to generate safe trajectories.
To represent the reference trajectories of the robot, this paper uses Bezier curves, $b: [0,1] \to \R$, which are defined as: 
\begin{equation} \label{eq:bez}
    b(t) = \sum_{v=0}^{V}b_vP_v(t),
\end{equation}
where $b_v\in\R$ are called the \emph{Bezier Coefficients} and $P_v: [0,1] \to \R$ is the \emph{Bernstein Basis Polynomial} \cite[(1)]{paper-bernstein} for each $v \in \{0,\ldots,V\}$.

The robot has $n$ joints in total, with $n_a$ actuated joints (motors) and $n_u$ unactuated (passive) joints, where $n_a + n_u = n$.
We denote the set of indices for all actuated joints as $\A\subset\N$ and the set of indices for all unactuated joints as $\U\subset\N$.
Thus, $\A \cap \U = \emptyset$ and $\A \cup \U = \{1,\ldots,n\}$.

We focus on robots with flat feet in this paper:
\begin{assum} \label{assum-flatfoot}
    The lower surface of the feet of the biped are flat rectangles.
\end{assum}
Motivated by \cite[Section~1.2]{paper-bipedalrobotdynamics}, the bipedal robot in this paper is modeled as a hybrid system that alternates between single-support and double-support phases depending on the number of legs in contact with the ground. 
In the single-support phase, the leg in contact with the ground is called the \emph{stance leg}, and the other leg is called the \emph{swing leg}.

\subsubsection{Single-Support Dynamics}
Suppose the duration of a single-support phase is $T$.
We define the trajectory of the robot as $q:[0,T]\rightarrow\Q$ where $\Q\subset\R^n$ is the configuration space of all joints of the robot.
We define $\qu(t)\in\R^{n_u}$ as the collection of all unactuated entries of $q(t)$ and $\qa(t)\in\R^{n_a}$ as the collection of all actuated entries of $q(t)$.
The constraint function is denoted by $c:\Q\rightarrow\R^{n_c}$ where $n_c\geq0$ is the number of constraints.
The constraints describe the closed-loop kinematic chain constraints and the requirement that the stance foot is fixed at a specific position and orientation.
The constraints are satisfied when $c(q(t)) = \mathbb{0}_{n_c}$.
The Jacobian of $c$ at $q(t)$ is denoted by $J(q(t))\in\R^{n_c\times n}$.
 
The robot's dynamics during the single-support phase for all $t\in[0,T]$ can be described as \cite[Section 8.1]{paper-featherstone}:
\begin{multline}
H(q(t))\ddq(t) + C(q(t),\dq(t))\dq(t) + g(q(t)) = \\
= \tau(t) + J^T(q(t))\lambda(t),
\end{multline}
\begin{align}
    &c(q(t)) = \mathbb{0}_{n_c} \label{eq:cons-pos} \\
    &\dot{c}(q(t)) = J(q(t))\dq(t) = \mathbb{0}_{n_c} \label{eq:cons-vel} \\
    &\ddot{c}(q(t)) = J(q(t))\ddq(t) + \dot{J}(q(t))\dq(t) = \mathbb{0}_{n_c}, \label{eq:cons-acc}
\end{align}
where $H(q(t)) \in \R^{n\times n}$ is the positive definite inertia matrix,
$C(q(t),\dq(t))\in \R^{n\times n}$ is the Coriolis matrix, 
$g(q(t)) \in \R^{n}$ is the gravitational force vector, 
and $\lambda(t)\in\R^{n_c}$ is the reaction forces/wrenches to maintain the constraints at time $t$.
The torque applied on each joint at time $t$, $\tau(t) \in \mathbb{R}^n$, is given by
\begin{equation}
    \tau(t) = Bu(t),
\end{equation}
where $u(t)\in\R^{n_a}$ is the control input and $B\in\R^{n\times n_a}$ is the transmission matrix, which follows the property that
\begin{equation}
    B_u = \mathbb{0}_{n_u\times n_a},\quad 
    B_a = \mathbb{1}_{n_a\times n_a},
\end{equation}
where $B_a$ and $B_u$ is the collection of all actuated rows and all unactuated rows of $B$, respectively.

The conditions for maintaining surface contact between the stance foot and the ground are given in \cite[(6)-(8)]{paper-trajopt-collocation}.
In particular, the reaction force/wrench to maintain all constraints at time $t$, $\lambda(t)\in\R^{n_c}$, includes force/wrench on unactuated joints to maintain closed-loop constraints, or force/wrench from the ground to maintain contacts between the stance foot and the ground.
Let $\lambda_{st}(t)$ be the 6-dimensional vector that is a collection of all entries that correspond to the contact constraints in $\lambda(t)$ (i.e., the contact wrench). 
The corresponding contact wrenches consist of three constraint forces, $(\lambda_{st}^{fx}(t), \lambda_{st}^{fy}(t), \lambda_{st}^{fz}(t))$, 
and three constraint moments, $(\lambda_{st}^{mx}(t), \lambda_{st}^{my}(t), \lambda_{st}^{mz}(t))$, respectively. 
To ensure that contact is maintained, the following conditions on contact wrenches must be satisfied:
\begin{subequations} \label{eq-contact-constraints}
\begin{align}
    0 &\leq \lambda_{st}^{fz}(t) \label{eq-contact-constraints-positive} \\
    \sqrt{(\lambda_{st}^{fx}(t))^2 + (\lambda_{st}^{fy}(t))^2} &\leq \mu \lambda_{st}^{fz}(t) \label{eq-contact-constraints-translation} \\
    \lambda_{st}^{mz}(t) &\leq \gamma \lambda_{st}^{fz}(t) \label{eq-contact-constraints-rotation} \\
    -\frac{1}{2}l_a\lambda_{st}^{fz}(t) \leq \lambda_{st}^{mx}(t) &\leq \frac{1}{2}l_a\lambda_{st}^{fz}(t) \label{eq-contact-constraints-flipx} \\
    -\frac{1}{2}l_b\lambda_{st}^{fz}(t) \leq \lambda_{st}^{my}(t) &\leq \frac{1}{2}l_b\lambda_{st}^{fz}(t) \label{eq-contact-constraints-flipy},
\end{align}
\end{subequations}
where $\mu$ is the translational friction coefficient of the ground, 
$\gamma$ is the torsional friction coefficient of the ground,
$l_a$ and $l_b$ are the length and width of the lower surface of the stance foot.
Note these conditions correspond to the following condition:
the ground reaction force should be non-negative \eqref{eq-contact-constraints-positive}, 
the stance foot should not slide on the ground in translation direction \eqref{eq-contact-constraints-translation}, or in rotation direction \eqref{eq-contact-constraints-rotation}, and the robot should not roll over the edge of the stance foot \eqref{eq-contact-constraints-flipx}, \eqref{eq-contact-constraints-flipy}. 

\subsubsection{Double Support Dynamics}
To simplify the exposition and formulation of the real-time optimization problem, we follow the assumption made in \cite[Section 3.2, HGW3]{paper-bipedalrobotdynamics}:
\begin{assum} \label{ass-double-support} 
The double support phase is instantaneous, and the associated impact due to ground contact is modeled as a rigid contact. 
\end{assum}
We describe the instantaneous change in the robot model caused by the impact using the notion of a guard and reset map as in the definition of hybrid systems \cite[Definition 7]{paper-burden2015metrization}.
The force of ground contact imposes a holonomic constraint on the position of the stance foot that enables one to construct a reset map \cite[Section 3.4.2]{paper-bipedalrobotdynamics}:
\begin{equation} \label{eq-resetmap}
(q^+,\dq^+,\lambda_r)=\Delta(q^-,\dq^-),
\end{equation}
where $\Delta$ describes the relationship between the pre-impact joint angles, $q^-$, and velocities, $\dq^-$, and post-impact joint angles, $q^+$, velocities, $\dq^+$, and reaction force/wrench $\lambda_r$. 
For brevity, we do not include an explicit formula for $\Delta$ in this paper, but it can be found in \cite[(3.20)]{paper-bipedalrobotdynamics}.
Note that this instantaneous change only affects the velocities of the robot joints (i.e., $q^+=q^-$).

\subsubsection{Fully-Actuated Representation}

When the number of constraints $n_c$ is equal to the number of unactuated dimensions $n_u$, the system is fully actuated.
This case usually holds for bipedal robots with actuated ankles and flat feet in the single-support phase \cite[Section 10]{paper-bipedalrobotdynamics}.
More details on the discussion on Digit and other similar robots can be found in Section \ref{sec:simulation}.
The fully-actuated representation facilitates the control and optimization of such robots because it enables us to describe the unactuated joints trajectory as a function of the actuated joints trajectory.

Before formally describing this theorem, we make the following assumption:
\begin{assum} \label{assum-uniqueIK}
    During the single-support phase, for $\forall\qa(t)\in\Q_a$, there exists one and only one $\qu(t)\in\Q_u$ such that $c(q(t)) = \mathbb{0}_{n_c}$.
\end{assum}
\noindent This assumption ensures that the constraint Jacobian $J_u(q(t)) \in \R^{n_c\times n_u}$, 
which is the collection of unactuated columns of the constraint Jacobian $J(q(t))$ and a square matrix when $n_u = n_c$, 
is always invertible according to the inverse function theorem \cite[Theorem 5.2.1]{paper-inverse-function-theorem}.
Note in the experimental section, we numerically evaluate whether this assumption is satisfied.
We denote the $\qu$ that satisfies the constraints as a function of $\qa$:
\begin{equation} \label{eq-IK}
    \qu = \Gamma(\qa).
\end{equation}
Under this assumption, one can prove that the control input can be uniquely computed as a function of the actuated joint position, velocity, and acceleration.
The proof can be found in our online supplementary material \cite[Appendix I]{paper-appendix}.
\begin{thm} \label{thm-fully-actuated}
    Given actuated joint position $\qa(t)$, velocity $\dqa(t)$, and acceleration $\ddqa(t)$, the velocity and acceleration of all joints are given as
    \begin{align} 
        &\dq(t) = G(q(t))\dqa(t) \label{eq-fill-full-joints1} \\
        &\ddq(t) = G(q(t))\ddqa(t) + \dot{G}(q(t))\dqa(t). \label{eq-fill-full-joints2}
    \end{align}
    The reaction force $\lambda(t)$ and the control input $u(t)$ are then uniquely given as
    \begin{align} 
        \lambda(t) &= (J_u^{-1}(q(t)))^T\Tilde{\tau}_u(t) \label{eq-constrainedID-lambda} \\
        u(t) &= \Tilde{\tau}_a(t) - J_a(q(t))^T\lambda(t) = G^T(q(t))\Tilde{\tau}(t), \label{eq-constrainedID-torque}
    \end{align}
    where $\Tilde{\tau}(t)$ is the full inverse dynamics vector:
    \begin{equation}
        \Tilde{\tau}(t) = H(q(t))\ddq(t) + C(q(t),\dq(t))\dq(t) + g(q(t)).
    \end{equation}
    $\Tilde{\tau}_u(t)$ and $\Tilde{\tau}_a(t)$ are the collection of unactuated and actuated entries of $\Tilde{\tau}(t)$, respectively,
    $J_u(q(t))$ and $J_a(q(t))$ are the collection of unactuated and actuated columns of the constraint Jacobian $J(q(t))$, respectively,
    $G(q(t))\in\R^{n\times n_a}$ is the projection matrix from actuated space to joint space with its unactuated rows defined as
    \begin{equation}
        G_u(q(t)) = -J_u^{-1}(q(t)) J_a(q(t)),
    \end{equation}
    and its actuated rows defined as
    \begin{equation}
        G_a(q(t)) = \mathbb{1}_{n_a \times n_a}.
    \end{equation}
\end{thm}
\noindent As we show in the next section, this theorem allows us to formulate the trajectory optimization problem with decision variables only corresponding to the actuated joints. 
In particular, we can apply this theorem to recover the remaining joints, reaction force, and the control input as a function of just the actuated joints.
\section{Multiple-step Periodic Gait Generation}
\label{sec:gaitgeneration}

This section describes the optimization formulation for offline generating a library of desired gaits.

The duration of all walking steps to be a fixed number $T\in\R_+$.
The trajectory of the actuated joint $j\in\A$ for one walking step is a $V$-degree Bezier curve:
\begin{equation} \label{eq-gait-bezier-curve}
    \qGaitj(t) = \sum_{v = 0}^Vb_{v,j}P_v\left(\frac{t}{T}\right),
\end{equation}
where $t\in[0,T]$.
For an $L$-step gait optimization, the variables at a walking step $l \in \{1,\ldots,L\}$ are denoted by $b_{v,j}^{(l)}$.
The following variables are decision variables in the optimization problem: $\{b_{v,j}^{(l) \in \R}\}_{v\in\{0,\ldots,m\},j \in \A,l\in \{1,\ldots,L\}}$,
$\{\dq_{r}^{(l)}\in\R^n\}_{l \in \{1,\ldots,L\}}$, and $\{\lambda_{r}^{(l)}\in\R^{n_u}\}_{l \in \{1,\ldots,L\}}$,
where $\dq_{r}^{(l)}$ denotes the joint velocity after the reset map at the end of each step $l$, and
$\lambda_{r}^{(l)}$ denotes the reaction force during the reset map at the end of each step $l$ that is required to maintain the corresponding constraints.
The variable $y$ describes the concatenation of all of the variables above.

To ensure that the computed trajectory is dynamically feasible, the optimization problem checks that certain constraints are satisfied by the trajectory on a finite number of time nodes. 
Motivated by \cite{paper-chebyshev}, we choose the \defemph{Chebyshev Nodes}, $t_i \in [0,T]$  for $i \in \{1,\ldots,N\}$ as the set of time nodes along which constraints must be satisfied
where $N\geq3$ is the total number of the nodes.
This is then used to formulate the optimization problem to design desired gaits:
\begin{subequations} 
\begin{align} 
    \min_{y}&\quad\J(y) \label{eq-offlineopt-cost}\\
    \text{s.t.}&\quad \text{Joint Limits}\ \label{eq-offlineopt-con-jointlimits} \\
    & \quad \text{Actuator Limits} \label{eq-offlineopt-con-torquelimits} \\
    & \quad \text{Maintaining Contact} \label{eq-offlineopt-con-contact} \\
    & \quad \text{Torso Constraints} \label{eq-offlineopt-con-torso} \\
    & \quad \text{Swing Foot Constraints} \label{eq-offlineopt-con-swingfoot} \\
    & \quad \text{Reset Map Constraints} \label{eq-offlineopt-con-resetmap}
\end{align}
\end{subequations}
The remainder of this subsection describes the cost and each of the constraints.
Before proceeding, recall that one can use the decision variables of the optimization problem to compute the control input \cite[Theorem 2]{zhang2024system}.

\subsubsection{Cost \eqref{eq-offlineopt-cost}}
Inspired by \cite[(6)]{paper-centriodal}, the cost function aims to minimize the 2-norm of the control input consumed, the initial velocity, and the initial acceleration: 
\begin{multline} \label{eq:cost}
    \J(y) = \sum_{l=1}^{L}(\frac{w_1}{N} \sum_{i=1}^N \norm{u^{(l)}_i}_2 + \\
    + w_2 \norm{\dqGaita^{(l)}(0)}_2 + w_3 \norm{\ddqGaita^{(l)}(0)}_2),
\end{multline}
where $w_1$, $w_2$ and $w_3$ are user-defined positive constant scalars.

\subsubsection{Joints Limits \eqref{eq-offlineopt-con-jointlimits} \& Actuator Limits \eqref{eq-offlineopt-con-torquelimits}}
All joints $\qGait^{(l)}(t_i)$ belong to the configuration space $\Q$ for all $i \in \{1,\ldots,N\}$ and $l\in\{1,\ldots,L\}$.
The control input  $u^{(l)}(t_i)$ must be within the robot torque limits for all $i\in\{1,\ldots,N\}$ and $l\in\{1,\ldots,L\}$.

\subsubsection{Maintaining Contact \eqref{eq-offlineopt-con-contact}}
The contact constraints are formulated for  
$\lambda^{(l)}(t_i)$ and $\lambda^{(l)}_r$ for all $i\in\{1,\ldots,N\}$ and $l\in\{1,\ldots,L\}$ using \eqref{eq-contact-constraints} and can be computed using the decision variables by applying \eqref{eq-constrainedID-lambda}.

\subsubsection{Torso Constraints \eqref{eq-offlineopt-con-torso}}
Inspired by \cite[(50)]{paper-trajopt-collocation}, the torso constraints require that the absolute value of the roll and pitch angle of the torso are less than or equal to $3^{\circ}$ and the height of the torso larger than a specific value to avoid falling.

\subsubsection{Swing Foot Constraints \eqref{eq-offlineopt-con-swingfoot}} 

The swing foot constraints require that: the swing foot stays on the ground at the beginning and end of each step, the swing foot yaw angle is equal to a user specified desired angle at the beginning and end of each trajectory, the swing foot height is greater than or equal to some user specified height during the middle of the step, and that the swing foot position is equal to some user specified desired position at the beginning and at the end of every step. 
More information on the user-defined values can be found in Section \ref{sec:simulation}.

\subsubsection{Reset Map Constraints \eqref{eq-offlineopt-con-resetmap}}
Reset map constraints require that the previous walking step and the next walking step are ``aligned" according to \eqref{eq-resetmap}:
\begin{subequations}
    \begin{align}
        &(q^{(l)}(x_N), \dq_{r}^{(l)}, \lambda_{r}^{(l)}) = \Delta(q^{(l)}(x_N), \dq^{(l)}(x_N)) \label{eq-resetmapcons}\\
        &\qGaita^{(l+1)}(x_1) = \qGaita^{(l)}(x_N) \label{eq-resetmapcons-pos} \\
        &\dqGaita^{(l+1)}(x_1) = \dq_{ra}^{(l)} \label{eq-resetmapcons-vel}
    \end{align}
\end{subequations}
for all $l \in \{1,\ldots, L-1\}$.
\eqref{eq-resetmapcons-pos} ensures that the actuated joint positions are continuous between steps.
\eqref{eq-resetmapcons} and \eqref{eq-resetmapcons-vel} ensure that the actuated joint velocities satisfy the reset map equation \eqref{eq-resetmap} between steps.

For periodic gaits, we generally assume that the number of walking steps is even,
so that the stance legs at the first step and the last step are different.
The corresponding reset map constraints can be formulated as below:
\begin{subequations}
    \begin{align}
        &(q^{(L)}(x_N), \dq_{r}^{(L)}, \lambda_{r}^{(L)}) = \Delta(q^{(L)}(x_N), \dq^{(L)}(x_N)) \\
        &\qGaita^{(1)}(x_1) = \qGaita^{(L)}(x_N) \\
        &\dqGaita^{(1)}(x_1) = \dq_{ra}^{(L)}
    \end{align}
\end{subequations}
If the number of walking steps is odd, in other words, the stance legs at the first step and the last step are the same, the constraints need to be rewritten so that joint positions and velocities at the beginning of the first step and the end of the last step are symmetric.
As a result, this only holds for symmetric walking behaviors, such as walking straight forward.

\section{Simulation Experiments}
\label{sec:simulation}

This section summarizes the experimental evaluation of our method on Digit and its comparison to Aligator \cite{paper-aligator}.
Note we have only compared to Aligator rather than other optimization-based trajectory design algorithms because Aligator (1) is able to natively use Pinocchio \cite{paper-pinocchio} which efficiently evaluate derivatives of a robot's dynamics, (2) has made its code publicly available, and (3) it is able to enforce closed-loop kinematic constraints.
Note, we also performed the same comparison on another humanoid Talos \cite{paper-talos} that was originally included in the official examples of aligator and does not contain any closed-loop kinematics chains.
The results can be found in our online supplementary material \cite{paper-appendix}.
We have also shown that our method works for other fully-actuated systems, such as robotic manipulators.
More results can be found on our project website \href{https://roahmlab.github.io/RAPTOR/}{https://roahmlab.github.io/RAPTOR/}.

\subsection{Digit Overview}
\label{sec:digit-overview}

\begin{figure}[t]
    \centering
    \includegraphics[width=0.8\columnwidth]{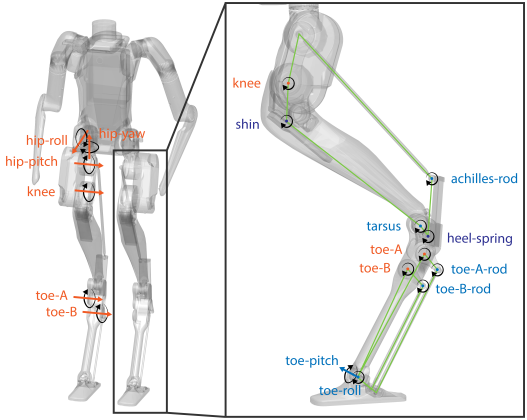}
    \caption{This figure illustrates the closed-loop structure on the legs of Digit. 
    The orange arrows show the rotation axes of all actuated joints (motors). 
    The blue arrows show the rotation axes of all unactuated joints. 
    The purple arrows show the rotation axes of all springs (shin and heel-spring), which are assumed to be fixed in this paper. 
    The green lines show how these joints are connected to be closed-loops.}
    \label{fig-digitv3}
\end{figure}

\defemph{Digit} is a humanoid robot with 42 degrees of freedom, developed by Agility Robotics (including floating-base).
Digit includes 4 actuated revolute joints per arm and 14 revolute joints per leg, of which 6 are actuated and 8 are passive.
Each leg has 3 closed kinematic chains. The upper leg chain consists of 1 actuated joint (knee) and 4 passive joints (shin, heel-spring, achilles-rod, tarsus). 
The lower leg chains include 2 actuated joints (toe-A, toe-B) and 4 passive joints (toe-pitch, toe-roll, toe-B-rod, toe-B-rod).
The kinematic structure is summarized in Figure \ref{fig-digitv3}.

We make the following simplifications for Digit.
We focus on the leg motion of Digit in this work.
In addition, note the springs on Digit's legs exhibit little movement during actual hardware experiments.
In fact, the most recent bipedal robot models (\cite{paper-atlas, paper-teslabot, paper-unitree, paper-fourier-intelligence}), including the latest version of Digit \cite{paper-agility-v4}, do not incorporate such springs.
As a result, we make the following assumption:
\begin{assum} \label{assum-arms}
    All arms on Digit are assumed to be fixed.
    All four springs (shin and heel-spring on both legs) on Digit are assumed to be fixed.
\end{assum}
    

By fixing all the springs, each chain in the legs becomes a four-bar linkage mechanism with 3 kinematics constraints.
This results in a total of 15 joints per leg with 9 passive joints and 9 constraints.
To numerically verify Assumption \ref{assum-uniqueIK}, we provide $100$ different actuated joint positions within the corresponding joint limits and solve inverse kinematics for unactuated joints.
The solutions are always unique.
The other unactuated joints, which are the floating-base coordinates, are also unique once the unactuated joints in the closed-loop are settled.
As a result, Digit can be treated as a fully actuated system during the single-support phase.

\subsection{Implementation Details}

\subsubsection{Platform Details}
Our gait optimization code is implemented in C++.
The experiments are run on a desktop with an AMD Ryzen 9 5950 16-core 32-thread processor and 128 GB RAM.
In our method, we use Pinocchio \cite{paper-pinocchio} to compute the unconstrained inverse dynamics and its gradient.
We use Ipopt \cite{paper-ipopt} as our nonlinear solver.
The first-order gradient is provided analytically while the second-order Hessian is approximated using Ipopt's "limited-memory" option.
We use linear solvers from HSL \cite{paper-hsl}.

\subsubsection{Handling Closed-loop Constraints}
To apply Theorem \ref{thm-fully-actuated}, we must first solve for the inverse kinematics $\Gamma$.
This enables the computation of the other terms in the theorem.
We use the multidimensional root finding function in the GSL library \cite{paper-gsl} to find $q_u(t)$.
This method is sensitive to its initial guess.
To be more specific, in the offline operation, we uniformly sample 100 points within the actuated joint limits and solve the inverse kinematics problem over each of these 100 points, as how we have numerically verified Assumption \ref{assum-uniqueIK} described at the end of Section \ref{sec:digit-overview}.
We fit a trigonometric series to approximate the solutions.
In that way, we can compute an approximate solution rapidly by relying upon interpolation.
During each iteration of the optimization, we treat this approximation as the initial guess of the GSL multidimensional root-finding function, which then converges into a more accurate solution, usually in less than 10 iterations.
More details on implementation can be found in the README\footnote{\label{digit-readme-footnote}\href{https://github.com/roahmlab/RAPTOR/tree/main/Examples/Digit}{https://github.com/roahmlab/RAPTOR/tree/main/Examples/Digit}} of our code.

\subsection{Gait Optimization Comparison}

\begin{figure*}[t]
    \centering
    \includegraphics[width=2.0\columnwidth]{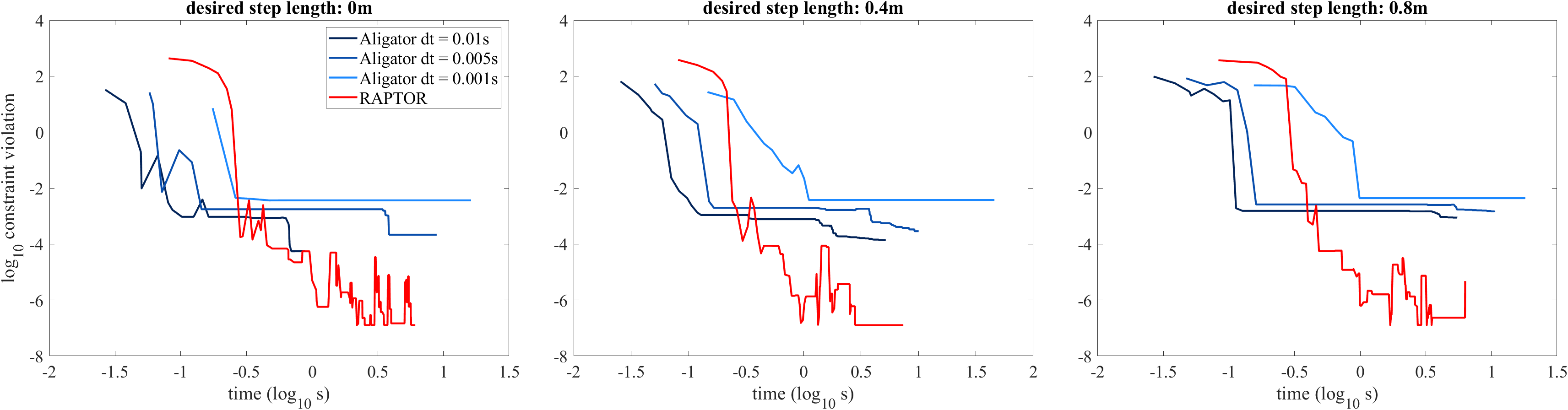}
    \caption{Pareto curve of the constraint violation with respect to the computation time. 
    Note that all curves do not start from 0 in time, since we need to evaluate the problem for at least one iteration to get the constraint violation.
    In other words, the beginning of the curves indicates the computation time of the first iteration.}
    \label{fig:solve-time-comparison-Digit}
\end{figure*}

We perform a comparison between RAPTOR and aligator \cite{paper-aligator} on Digit.
Our implementation on aligator has also been open-sourced\footnote{\href{https://github.com/roahmlab/aligator-roahmlab}{https://github.com/roahmlab/aligator-roahmlab}}.
We perform 3 different experiments which make Digit walk forward for 0m, 0.4m, and 0.8m forward in one walking step.
We assign a fixed initial configuration of the robot to aligator and enforce the same constraint for RAPTOR, so that both methods start from the same configuration at the beginning of the trajectory.
We set the maximum number of iterations to 200 and the constraint violation tolerance to 1e-4 for both methods.
For aligator, we adhere to the same settings as provided in the official example except the time discretization.
We consider 3 different time discretizations (dt = 0.01s, 0.005s, 0.001s) for further discussion on aligator.
For RAPTOR, we performed an ablation study to choose the best parameters for Ipopt.
The results can be found in our online supplementary material \cite[Appendix II]{paper-appendix}.
We choose the degree of the Bezier curve $V$ to 5 and the number of time nodes $N$ to 14.
We choose HSL ma57 as the linear solver and the update strategy for barrier parameter to \texttt{adaptive}.
To initialize the optimization, we adhere to the same initial guess strategy as provided in the official example for aligator.
For RAPTOR, we simply assign all decision variables to 0.
More details on warm up strategy of RAPTOR can be found in README\footnotemark[1] of our code.

To validate whether solution is physically realizable, we simulate the dynamics of the robot using the integration method \texttt{solve_ivp} from \cite{paper-scipy} in Python.
Since the development of the tracking controller is out of the scope of this work, we simply apply a naive control policy here:
\begin{equation} \label{eq-controller}
    u(t) = u_{open}(t) + K_P (q_d(t) - q(t)) + K_D (\dot{q}_d(t) - \dot{q}(t)),
\end{equation}
where $u_{open}(t)$ is the open loop control input generated from the optimization, $K_P$ and $K_D$ are the PD gains.
We abuse the notation here and denote $q(t)$ as the robot states in the simulation and $q_d(t)$ as the desired states of the optimized gait trajectory.
aligator generates optimized control inputs on its corresponding time discretization, hence we compute $u_{open}(t)$ using zero-order hold (ZOH) over the control input sequence.
The motivation of choosing different time discretization here is that, intuitively speaking, finer discretization should lead to better tracking performance when using ZOH.
For RAPTOR, since the outputs are continuous trajectories, we can directly compute $u_{open}(t)$ at any time $t$ using Theorem \ref{thm-fully-actuated}.
We tuned the PD gains to minimize the tracking error while staying within the torque limitations and eventually choose $K_P = 80$ and $K_D = 5$.

Figure \ref{fig:solve-time-comparison-Digit} shows how the constraint violation converges with respect to time during optimization.
Although aligator converges faster than RAPTOR in general, it struggles to converge to an acceptable solution for more challenging tasks (desired step length $\geq$ 0.4m) or for finer time discretization, while RAPTOR is always able to converge to a feasible solution below the constraint violation tolerance, usually within 1 second.

\begin{figure}[t]
    \centering
    \includegraphics[width=0.75\columnwidth]{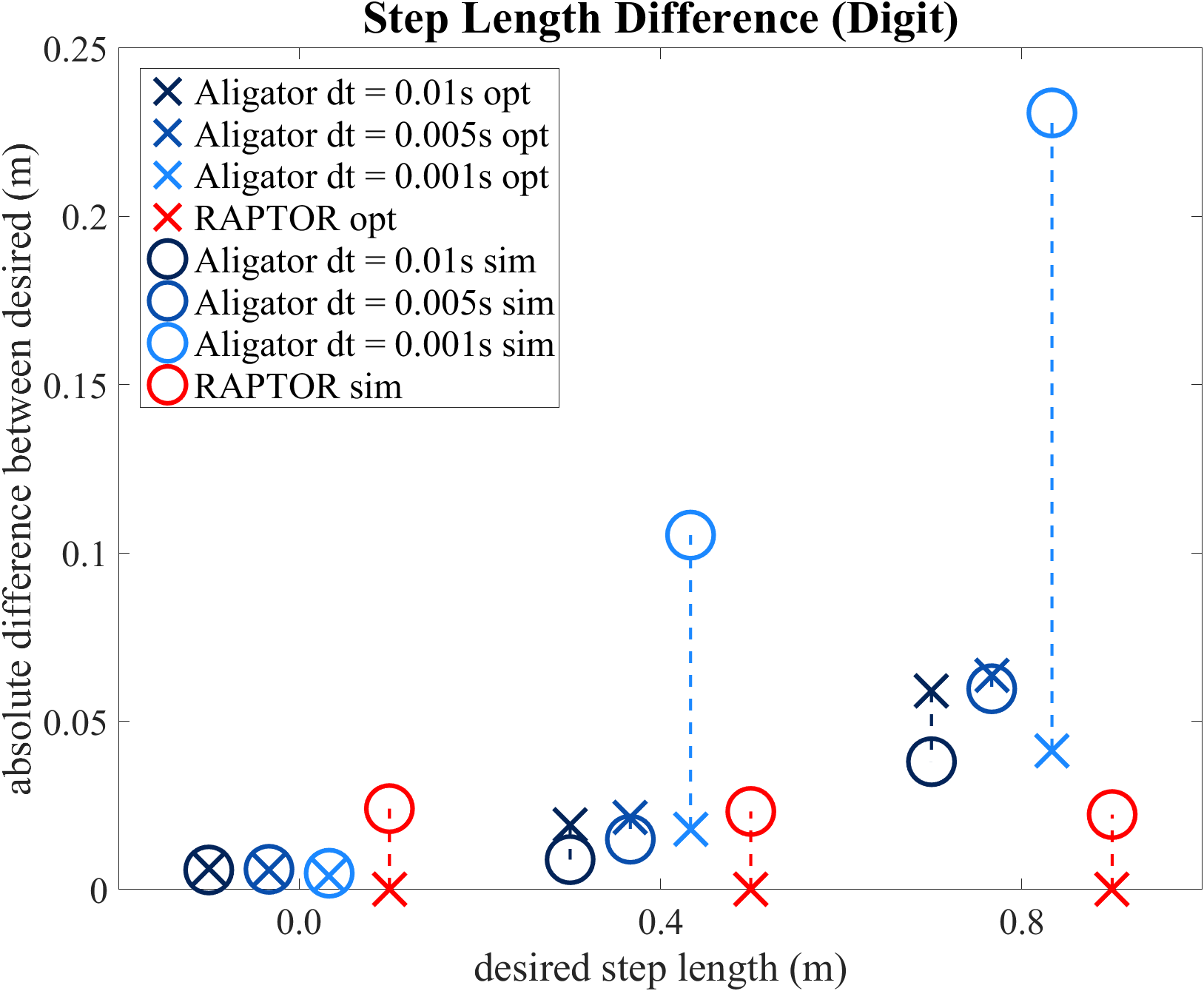}
    \caption{The absolute value of the difference between the desired step length and the actual step length.
    The crosses indicate the difference on the optimized trajectory.
    They may be away from 0 due to optimization not converging to a feasible solution.
    The circles indicate the difference on the trajectory in simulation when tracking the optimized trajectory using a controller.
    The circles may not be aligned with the crosses due to imperfect tracking of the controller.}
    \label{fig:swingfoot-task-comparison-Digit}
\end{figure}

\begin{figure}[t]
    \centering
    \includegraphics[width=0.75\columnwidth]{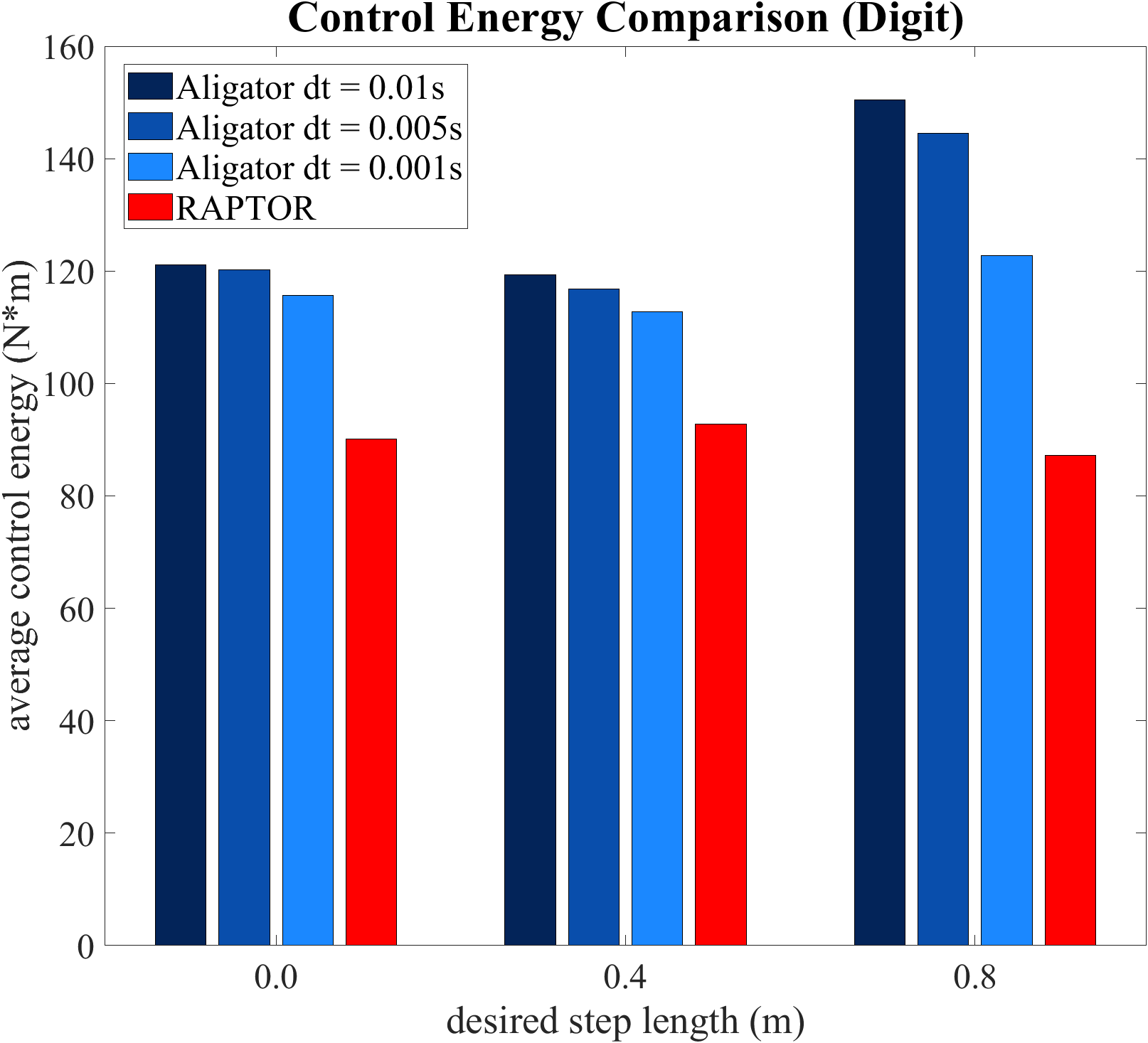}
    \caption{Control energy consumption of 3 variants of aligator and RAPTOR in the simulation for 3 different desired step lengths}
    \label{fig:control-energy-comparison-Digit}
\end{figure}

Figure \ref{fig:swingfoot-task-comparison-Digit} provides more details on the impacts of converging below the constraint violation tolerance or not.
Here we consider the difference between the desired step lengths and the resultant step lengths of both the optimized trajectories and the simulation trajectories when tracking the optimized trajectories using the controller in \eqref{eq-controller}.
As we have discussed previously, aligator struggles to generate a feasible solution that reaches the desired step length in the optimization stage, while the solution of RAPTOR can always reach desired step lengths.
For simulation trajectories, finer discretization brings better tracking performance for aligator when compared between 0.01s and 0.005s.
However, aligator struggles to converge when time discretization is 0.001s, returning unoptimized control inputs that lead to larger tracking error, and as a result performs the worst in both optimization and simulation.
For RAPTOR, \eqref{eq-controller} results in a larger tracking error compared to aligator. 
The performance is still better than that of aligator for the most challenging task when the desired step length is 0.8m.
We can infer that RAPTOR could perform better in simulation if equipped with a more sophisticated controller, which we consider to be one of our future work.

We also consider the control energy consumed in the simulation, which is calculated using $\sqrt{\frac{1}{N}\sum_{i=1}^N u_i^T u_i }$,
where $N$ is the number of samples recorded in the simulation and $u_i$ is the control input at $i$th time sample.
As shown in Figure \ref{fig:control-energy-comparison-Digit}, RAPTOR is always able to find the solution with the minimum control energy consumption, which is critical for humanoids with embedded mobile batteries.

\section{Conclusion}
\label{sec:conclusion}

This paper presents a novel trajectory optimization algorithm for full-size humanoids that can generate feasible gaits in seconds. 
The method outperforms existing state-of-the-art planners in terms of energy efficiency and achieving the desired behavior.
Future work will involve transferring the optimized trajectory onto real-world hardware.

\renewcommand{\bibfont}{\normalfont\footnotesize}
{\renewcommand{\markboth}[2]{}
\printbibliography}

\end{document}